%% file: main.tex
\documentclass[10pt,twocolumn,letterpaper]{article}

 \usepackage[pagenumbers]{cvpr} 
\usepackage{multirow}
\usepackage{soul}
\usepackage[accsupp]{axessibility}  


\definecolor{cvprblue}{rgb}{0.21,0.49,0.74}
\usepackage[pagebackref,breaklinks,colorlinks,allcolors=cvprblue]{hyperref}

\def\paperID{63} 
\def\confName{CVPR}
\def\confYear{2025}

\title{Shopformer: Transformer-Based Framework for Detecting Shoplifting via Human Pose
}

\author{
  Narges Rashvand \quad Ghazal Alinezhad Noghre \quad Armin Danesh Pazho\\ 
  Babak Rahimi Ardabili \quad Hamed Tabkhi\\
  University of North Carolina at Charlotte\\
  Charlotte, NC, USA\\
  {\tt\small \{nrashvan, galinezh, adaneshp, brahimia, htabkhiv\}@charlotte.edu}
}

\begin{document}
\maketitle
\input{sec/abstract}    
\input{sec/intro}

\input{sec/related_works}
\input{sec/shopformer}

\input{sec/experiments}

\input{sec/results}
\input{sec/conclusion}

{
    \small
    \bibliographystyle{ieeenat_fullname}
    \bibliography{main}
}


\input{sec/suppl}

\end{document}

%% file: sec/abstract.tex
\begin{abstract}

Shoplifting remains a costly issue for the retail sector, but traditional surveillance systems, which are mostly based on human monitoring, are still largely ineffective, with only about 2\% of shoplifters being arrested. Existing AI-based approaches rely on pixel-level video analysis which raises privacy concerns, is sensitive to environmental variations, and demands significant computational resources. To address these limitations, we introduce Shopformer, a novel transformer-based model that detects shoplifting by analyzing pose sequences rather than raw video. We propose a custom tokenization strategy that converts pose sequences into compact embeddings for efficient transformer processing. To the best of our knowledge, this is the first pose-sequence-based transformer model for shoplifting detection. Evaluated on real-world pose data, our method outperforms state-of-the-art anomaly detection models, offering a privacy-preserving, and scalable solution for real-time retail surveillance. The code base for this work is available at https://github.com/TeCSAR-UNCC/Shopformer.

\end{abstract}

%% file: sec/intro.tex
\section{Introduction}
\label{sec:intro}

\begin{figure}[htbp]
  \centering
   \includegraphics[trim=0pt 0pt 0pt 50pt, clip, width=3.2 in]{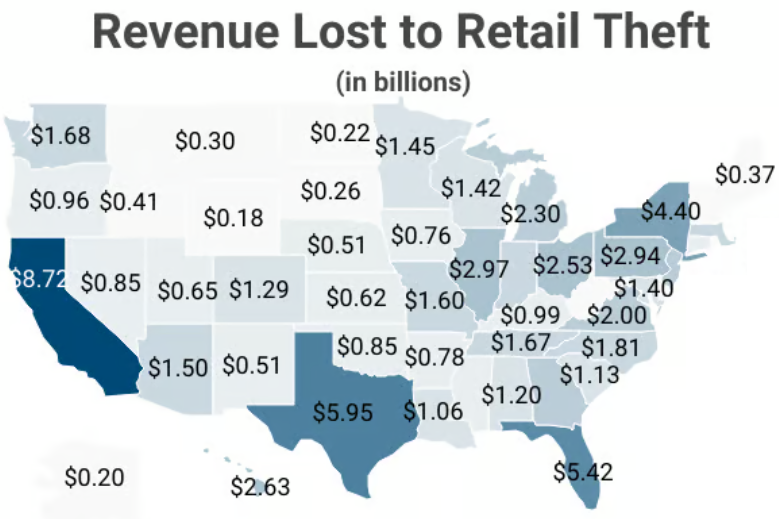}
   \caption{This map shows the revenue lost to retail shoplifting across U.S. states (in billions), highlighting the geographical distribution of shoplifting-related losses, and identifying the states experiencing the highest prevalence of retail shoplifting nationwide \cite{capitalone2025shoplifting}.}
   \label{fig:shoplifting_statistics}
\end{figure}

 Shoplifting is a growing problem that poses serious financial and operational challenges for retailers, impacting both businesses and the broader economy. Beyond financial damages, shoplifting leads to increased security costs, inventory issues, and higher prices for consumers \cite{ardabili2024exploring}. As illustrated in \cref{fig:shoplifting_statistics}, the map of retail shoplifting by U.S. states highlights the geographic distribution and economic burden of shoplifting nationwide, revealing the widespread scale of this issue \cite{capitalone2025shoplifting}. Despite these escalating losses, existing monitoring systems remain highly inefficient, as they rely heavily on manual surveillance. Even with these efforts, studies indicate that only about 2\% of shoplifters are apprehended \cite{ansari2022expert, rashvand2025exploring}, emphasizing the urgent need for more advanced, automated shoplifting detection methods to enhance real-time prevention.
 
\begin{figure}
\centering
\includegraphics[trim=128pt 140pt 81pt 115pt,clip, width= 3.2 in]{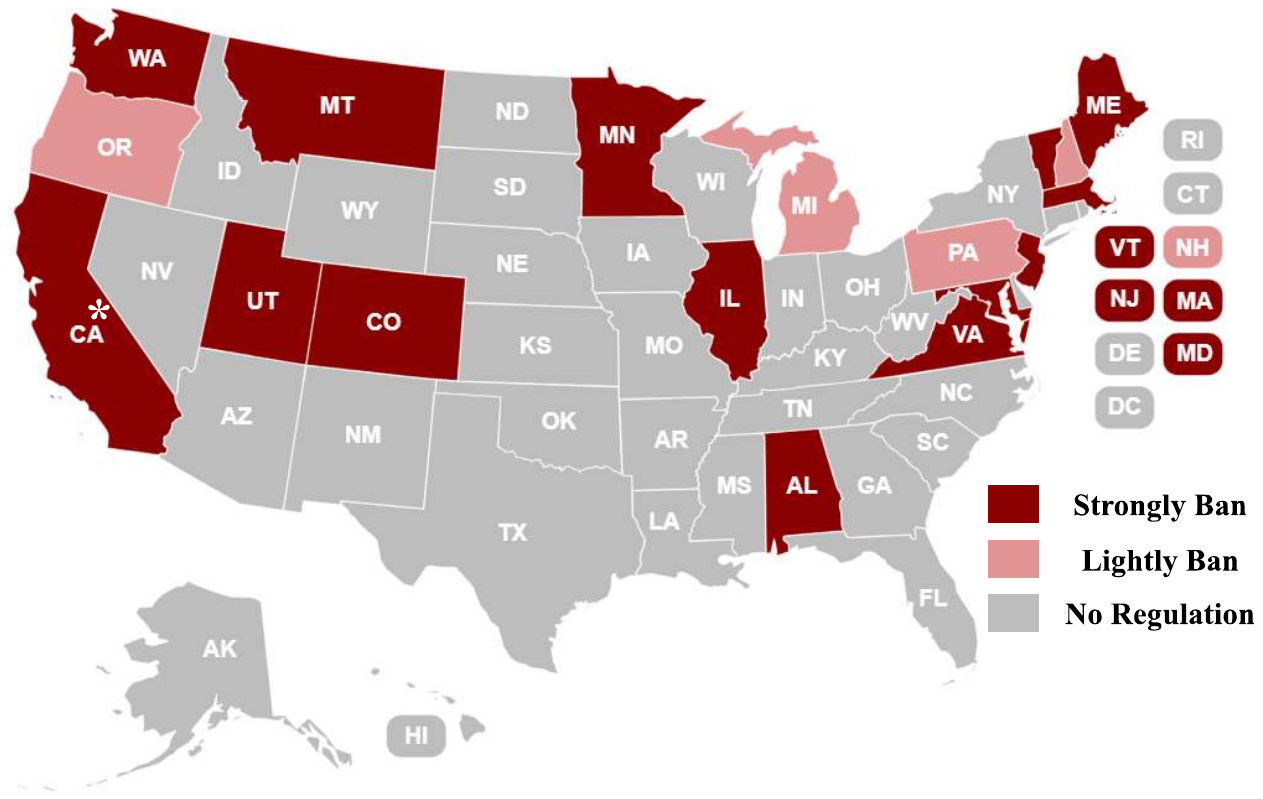}
\caption{Represents the status of facial recognition technology (FRT) laws in US states. Dark red states have strong limitations, while light red states have lighter limitations on the use of FRT. States without enacted state-level laws 
are shown in gray. \\ * Although California lacks state-level FRT regulations, some of its counties impose strict restrictions.}\label{FRT regulation map}
\end{figure}

 Interest in AI-powered detection systems is rapidly expanding across various domains \cite{babaey2025detecting, abshari2025survey, pazho2023survey}. In computer vision applications, AI-driven systems analyze video surveillance data in real-time to detect shoplifting, alert security personnel, and generate actionable insights to improve store safety and efficiency \cite{ansari2023optimized, kirichenko2022detection, arroyo2015expert, muneer2023shoplifting}. Recent research on shoplifting detection has focused on dataset creation and algorithm development, introducing novel approaches and benchmarking datasets \cite {gim2020automatic, ansari2023optimized, nazir2023suspicious, kirichenko2022detection, muneer2023shoplifting}. Despite growing research in AI-based shoplifting detection, significant challenges remain, as most approaches rely on pixel-level video analysis, making them highly sensitive to background variations, lighting changes, and occlusions common in real-world retail settings. Additionally, many approaches are trained on staged or simulated shoplifting scenarios, which fail to capture the complexity of real-life shoplifting behaviors. Analyzing entire video frames at each time step is also computationally intensive, making pixel-based methods less practical for real-time deployment in large-scale retail settings. Besides these practical limitations in using pixel-based algorithms, there are binding policies that ban the use of pixel-based algorithms in many applications \cite{ardabili2023understanding}. Although no specific regulation exists regarding the use of pixel-based algorithms, Facial Recognition Technologies (FRT), one of the most recognized pixel-based algorithms used in public safety applications, raising concerns about the violation of privacy-preserving policies \cite{ardabili2022understanding}. Figure \ref{FRT regulation map} illustrates how different US states regulate the use of FRT. As of December 2024, 13 states highlighted with dark red have enacted strong state-level regulation in limiting the use of FRT. Alabama, Illinois, Minnesota, Massachusetts, New Jersey, Vermont, Colorado, Maryland, Maine, Montana, Utah, Virginia, and Washington limits the use of FRT on serious crime cases and require warrant and notice. Although California lacks state-level policies, since the limitations in jurisdictions such as San Francisco is comparable to the other 13 states, we colored it in dark red. Oregon, Michigan, Pennsylvania, and New Hampshire in light red pose lighter restriction in this matter such as banning use of FRT in combination with police body cameras. According to available data from the U.S. Census Bureau, these 18 states account for nearly 43\% of the U.S. population \cite{uscensus}. While some states may have local-level regulations (e.g. using FRT in 18 major US airports including those in FL and TX), these states do not have prominent laws specifically regarding the use of FRT \cite{arnold2024introducing, latkowski2024facing, qandeel2024facial, wang2024beyond, gao2024frt, congress2023frt}.

While utilizing pixels and facial features is important aspects \cite{papaioannou2022mimicme, brooks2024deep, connolly2012adaptive, mathias2014face}, the outlined legal restrictions combined with the technical and societal implications of the shoplifting detection task, highlight the need to move beyond traditional pixel-level analysis and adopt pose-based approaches. By focusing on skeletal keypoints to represent human movement, pose-based algorithms 
eliminate irrelevant scene details and concentrate solely on human actions. This approach enhances robustness, reduces bias, ensures privacy preservation, and offers a scalable, computationally efficient solution for fair and ethical shoplifting detection. This further ensures that inherent societal biases would have a lower chance to impact the performance \cite{hu2025rethinking}.


Despite the advantages of pose-based approaches, no existing methods are specifically designed for shoplifting detection using pose dynamics. Instead, most prior works rely on pixel-level features, full-frame analysis, or consider shoplifting as a general anomaly detection task, overlooking its unique challenges.  

In this work, we address this gap by proposing Shopformer, a novel transformer-based model for shoplifting detection that directly processes pose sequences. Building on recent advancements in transformers for sequence modeling \cite{vaswani2017attention, noghre2024posewatch}, our approach utilizes human pose dynamics to identify shoplifting-related behaviors more effectively. A critical aspect of any transformer model is its tokenization process. To optimize this for shoplifting detection, we draw inspiration from graph convolutional networks \cite{yan2018spatial, chen2021channel, yu2017spatio, markovitz2020graph}, adapting and extending them to develop a specialized tokenization method. 
This strategy encodes pose sequences into compact, informative embeddings, enabling the transformer to focus on meaningful motion patterns while discarding irrelevant noise.
To the best of our knowledge, this is among the first works to explore transformer-based, pose-sequence-level models specifically designed for shoplifting detection, offering a new direction for behavior-centric, privacy-preserving, and real-time capable and scalable solutions in this domain. 

To demonstrate the effectiveness of our proposed model, we follow a strategy similar to \cite{rashvand2025exploring}, comparing it against state-of-the-art anomaly detection models. As no existing methods focus on pose-sequence-based shoplifting detection, this ensures a fair assessment of its effectiveness in capturing shoplifting-related behaviors. The key contributions of this study are:
\begin{itemize}
\item Introducing a Graph Convolutional Autoencoder (GCAE)-based tokenization 
method for converting raw pose data into compact, informative embeddings.
\item Introducing Shopformer, the first transformer-based model specially designed for shoplifting detection using human pose sequences, shifting from pixel-level analysis to a behavior-centric, privacy-preserving framework suitable for real-world retail environments.

\item A comprehensive evaluation of Shopformer on the only available real-world pose-based shoplifting dataset demonstrates its superior performance over three state-of-the-art pose-based anomaly detection models. 

\end {itemize}

%% file: sec/related_works.tex
\section{Related Works}
\label{sec:related works}
Existing research on shoplifting detection has primarily focused on two aspects: developing benchmark datasets and designing effective detection algorithms. In terms of dataset, several datasets have been introduced to advance shoplifting detection research, including those by \cite{arroyo2015expert, ansari2022expert, sultani2018real, muneer2023shoplifting}. These datasets are available in video format and are well-suited for developing pixel-based detection algorithms. However, they share critical limitations. With the exception of the UCF-Crime dataset \cite{sultani2018real}, which includes shoplifting as part of a broader collection of crime incidents sourced from YouTube videos, most datasets rely on staged scenarios where actors simulate theft using predefined actions. This controlled setting restricts their ability to capture the complexity and unpredictability of real-world shoplifting behaviors. Even though UCF-Crime includes real-world footage, it is not specifically designed for shoplifting detection and lacks detailed behavioral annotations and structured diversity necessary for developing pose-based shoplifting detection systems. Among existing shoplifting detection datasets, the recently introduced PoseLift dataset by \cite{rashvand2025exploring} is the only one that provides real-world pose sequences captured from actual retail environments. It focuses on pose-level data rather than raw videos and includes a diverse range of both shoplifting and normal behaviors.

On the algorithmic side, various pixel-based models have been proposed, often relying on and being evaluatrd on the UCF-crime dataset. For instance, Kirichenko et al. \cite {kirichenko2022detection} introduced a hybrid neural network that combines MobileNet for spatial feature extraction and GRUs for temporal sequence analysis. Similarly, Nazir et al. \cite{nazir2023suspicious} proposed a two-stage framework that integrates object tracking with time-series deep learning models, focusing on temporal features derived from bounding box trajectories rather than pixel-level features. Ansari et al. \cite{ansari2023optimized} presented a hybrid deep learning model that combines Inception V3 for spatial feature extraction with LSTMs for temporal sequence modeling, enabling the classification of video sequences as normal or shoplifting behaviors.


While these studies have provided valuable insights into shoplifting detection, several critical limitations remain unaddressed, as discussed in \cref{sec:intro}.
These challenges highlight the need for more practical, real-world solutions that focus on human behavior instead of raw pixel data. Pose-based analysis presents a promising approach to overcome these limitations. Given the absence of existing pose-sequence-based models for shoplifting detection, we explore four state-of-the-art anomaly detection models that utilize the pose sequences. One such approach is the Graph Embedded Pose Clustering (GEPC) model \cite{markovitz2020graph}, which detects anomalies by analyzing spatio-temporal pose graphs built from human keypoints. These graphs are encoded using a spatio-temporal graph convolutional autoencoder (ST-GCAE). The encoded features are then assigned to clusters using a deep clustering layer. To model normal behaviors, GEPC applies a Dirichlet Process Mixture Model (DPMM) on the soft assignment vectors, learning the distribution of normal actions in an unsupervised way. Another pose-based anomaly detection model, proposed by \cite{hirschorn2023normalizing}, also adopts an unsupervised approach, utilizing normalizing flows and spatio-temporal graph convolution blocks for human pose analysis.
TSGAD \cite{noghre2024exploratory}, a pose-based video anomaly detection (VAD) framework that integrates Graph Attentive Variational Autoencoders (GA-VAE) with trajectory prediction \cite{pazho2024vt, salzmann2020trajectron++, alinezhad2023pishgu} for detecting human-centric anomalies in an unsupervised manner. Another method is SPARTA \cite{noghre2024posewatch}, a transformer-based framework for anomaly detection. It introduces a Spatio-Temporal Pose and Relative Pose (ST-PRP) tokenization strategy and employs a Unified Encoder Twin Decoders (UETD) transformer architecture.

While these pose-based anomaly detection models have advanced human motion analysis, they remain primarily focused on general anomaly detection rather than shoplifting-specific behaviors. Unlike previous methodologies, our approach introduces a novel tokenization framework combined with a specially designed transformer architecture for shoplifting detection. Our model leverages a graph convolutional autoencoder to generate rich pose embeddings, which are then processed by an encoder-decoder transformer to effectively identify shoplifting behaviors.

%% file: sec/shopformer.tex
\section{Shopformer}
\label{sec:Shopformer}
As illustrated in \cref{fig:shopformer}, the Shopformer architecture is composed of two key components: the tokenizer module and the transformer module.
The architecture follows a two-stage training strategy. In the first stage, a Graph Convolutional Autoencoder (GCAE) is trained to learn meaningful representations of human pose sequences. The encoder of this trained GCAE is then frozen and integrated into the tokenizer module, as detailed in \cref{tokenizer}. In the second stage, the frozen encoder is used to generate tokens by converting raw pose data into compact, low-dimensional embeddings. These tokens are then passed to the transformer module, which captures temporal dependencies across the token sequence using self-attention mechanisms to enable shoplifting detection, as explained in \cref {transformer}.
\subsection{Tokenizer Module Based on GCAE Encoder}
\label{tokenizer}

 The input to Shopformer consists of pose sequences, as shown in \cref{fig:shopformer}, where each frame contains keypoint coordinates corresponding to an individual. The pose at a given time step $t$ is represented as:
\begin{equation}
P_t^i = [(x_1, y_1), (x_2, y_2), \dots, (x_N, y_N)]
\end{equation}
where $t$ represents the frame index, $i$ denotes the individual in the scene, $N$ is the number of keypoints, and each keypoint is defined by its (x, y) coordinates, indicating its position in the frame. Thus, the pose sequence for an individual over $n$
frames serves as the fundamental input to the model, capturing the spatial positions of the person's joints in each frame. This pose sequence is represented as:
\begin{equation}
PS_i = \{ P_t^i \mid t = t_0, t_0 + 1, \dots, t_0 + n - 1 \}
\end{equation}
where \( t_0 \) is the starting frame index of the sequence.

 Since the effectiveness of the transformer's self-attention mechanism depends heavily on how the input data is tokenized, and because attention operates at the token level, the quality of the tokenization process is critical. The tokenizer module is therefore responsible for converting raw human pose sequences into a structured and informative tokenized representation, ensuring that both spatial and temporal dynamics of human motion are effectively captured.

To fully leverage the power of self-attention for our downstream task of shoplifting detection, we employ a Graph Convolutional Autoencoder (GCAE) as the core of our tokenization strategy. Unlike traditional tokenization methods that rely on raw pose sequence data, such as the approach used in \cite{noghre2024posewatch}, the GCAE encoder-based tokenizer module embeds higher-order features into a structured latent space, making it well-suited for understanding human actions. This tokenizer is specifically designed to extract rich spatio-temporal features from human pose data using a graph-based representation. In our formulation, human pose is represented as a spatio-temporal graph, where each skeleton frame is modeled as a graph with nodes corresponding to keypoints (joints) and edges capturing both physical body connections and learned motion dependencies. This structure enables effective modeling of human motion over time.

Our tokenizer module is based on the encoder of a deep autoencoder architecture composed of Spatial-Temporal Graph Convolutional Network (ST-GCN) blocks \cite{yu2017spatio}. While prior works have used ST-GCN directly for classification or detection tasks, we instead leverage these blocks inside an autoencoder to learn compact and meaningful representations of pose sequences. Inspired by improvements in ST-GCN design \cite{markovitz2020graph}, which introduced more advanced spatial attention mechanisms and multiple GCN layers to model physical, dataset-level, and instance-specific relations, we adopt a symmetric architecture with layers of modified ST-GCN blocks for both the encoder and decoder. The encoder \( \mathcal{E}(\cdot) \) consists of stacked modified ST-GCN layers that map the pose input sequence \( {PS}_i \) into a lower-dimensional token sequence.
The autoencoder first applies spatial Graph Convolutional Networks (GCNs) to capture spatial relationships between keypoints. To capture temporal dynamics, the tokenizer also integrates Temporal Convolutional Networks (TCNs), which model motion patterns over time. The full GCAE is trained in an unsupervised fashion to reconstruct the original sequence. \begin{equation}
\hat{PS_i} = \mathcal{D}(\mathcal{E}(PS_i) 
\end{equation}
where \( \mathcal{E}(\cdot) \) and \( \mathcal{D}(\cdot) \) denote the encoder and decoder of the autoencoder, respectively. After training, only the encoder \( \mathcal{E} \) is retained and repurposed as the tokenizer module.

This design allows the trained encoder to generate spatio-temporal pose embeddings for the transformer module, which processes them for the shoplifting detection task.

\subsection{Transformer Module}
\label{transformer}
Following token generation with the GCAE-based tokenizer, the transformer becomes the core component of our framework. 

\begin{figure*}[htbp]
  \centering
   \includegraphics[trim=170pt 495pt 275pt 480pt, clip, width=6.9 in]{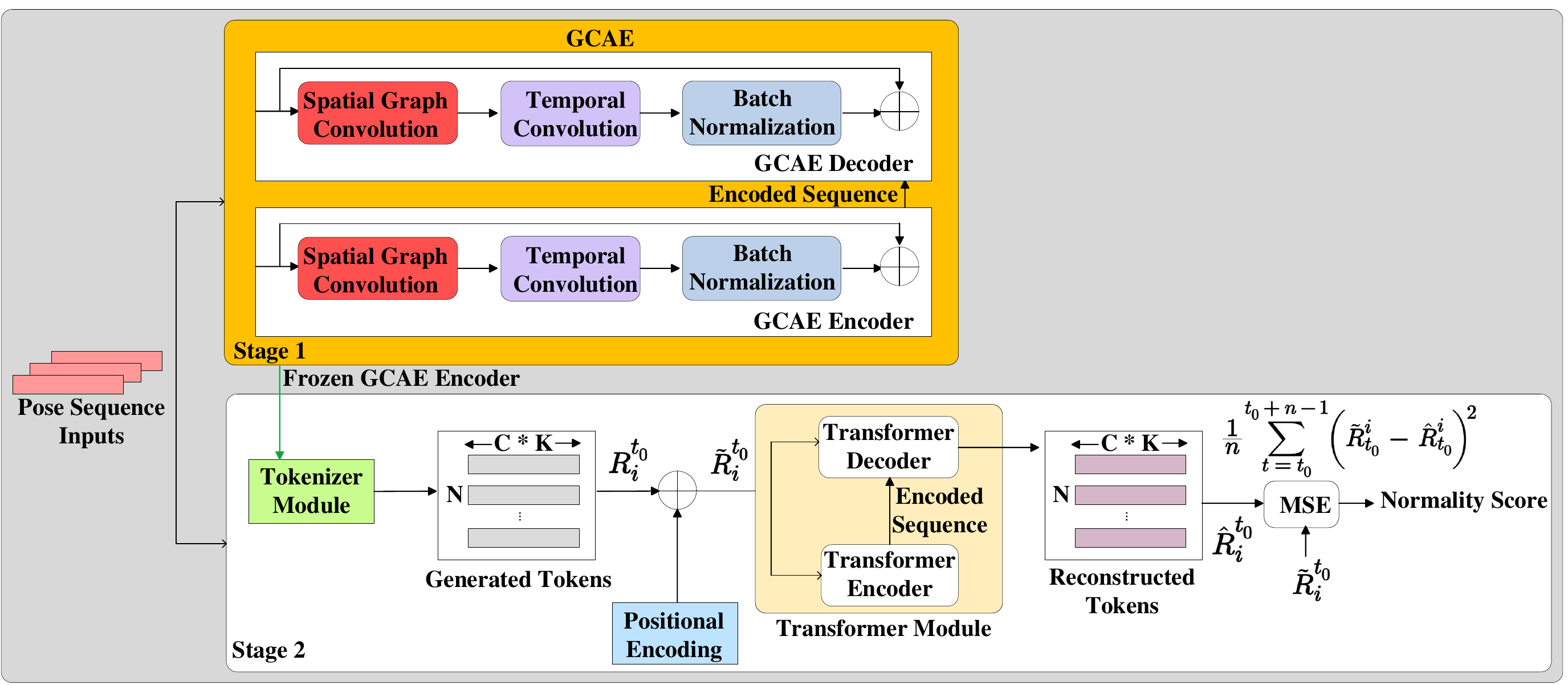}
   \caption{Overview of the Shopformer architecture. The framework operates in two stages: (1) a Graph Convolutional Autoencoder is first trained on pose sequences to learn rich spatio-temporal representations; (2) the pretrained encoder is then repurposed as a tokenizer module, generating compact tokens from input pose data. These tokens are passed through a transformer encoder-decoder module, which reconstructs the input sequence. The reconstruction error (MSE loss) is used to compute the normality score for shoplifting detection.}
   \label{fig:shopformer}
\end{figure*}

Inspired by SPARTA \cite{noghre2024posewatch}, which employs both Current Target Decoder (CTD) and  Future Target Decoder (FTD) branches, we propose
a simplified reconstruction-based strategy to model normal pose patterns.
This approach not only reduces architectural complexity and computational overhead but also results in a more lightweight and efficient transformer module, while still maintaining strong performance in shoplifting detection.

The transformer module follows the standard transformer architecture introduced in \cite{vaswani2017attention}, which is built entirely on self-attention mechanisms, compromising \(L\)   encoder-decoder layers, each equipped with \(T\) attention heads and a feed-forward network of dimensionality \(F\).
This design enables the model to efficiently capture long-range dependencies and process sequences in parallel, making it particularly effective for analyzing human movement patterns in real-time applications such as shoplifting detection. Our transformer module adopts an encoder-decoder architecture designed to process, learn, and reconstruct tokenized representations of pose sequences. The encoder maps the input tokens into a latent space, capturing meaningful representations, while the decoder reconstructs the token sequence from the latent features.

The token sequence ${R}^{t_0}_i$, generated by the tokenizer module and treated as a stream of sequential data, is defined  as:

\begin{equation}
R^{t_0}_i = [\text{Token}^{0}_i, \text{Token}^{1}_i, \dots, \text{Token}^{n-1}_i]
\end{equation}
Each token ${Token}^{k}_i$ represents a compact, spatio-temporal representation derived from the human pose. 

As illustrated in \cref{fig:shopformer}, the GCAE encoder-based tokenization strategy produces a sequence of \(N\)
tokens, where each token has a dimensionality of \(C \times K\), with \(C\) denoting the number of output channels from the GCAE encoder and \(K\) representing the number of keypoints.
Through extensive experimentation with different values of \(N\), as discussed in \cref {sec:token} and the supplementary materials, we identified the optimal token count \(N\) for our architecture. 

Since transformers do not inherently encode positional information, and tokenization itself does not preserve temporal order, we apply a positional encoding scheme \cite{vaswani2017attention}
to the token sequence \(R^{t_0}_i\) to incorporate sequence information into the input tokens. We denote the resulting positionally encoded sequence as \(
\tilde{R}^{t_0}_i
\), which 
serves as the input to the transformer module. 
This enriched sequence retains the semantic content of each token while embedding its temporal location within the sequence. The transformer module first encodes it into a latent representation and then decodes it to reconstruct the original token sequence, resulting in the output sequence:

\begin{equation}
\hat{R}^{t_0}_i = [\text{Token}'^{0}_i, \text{Token}'^{1}_i, \dots, \text{Token}'^{n-1}_i]
\end{equation}
where $\hat{R}^{t_0}_i$ is the reconstructed token sequence, and $\text{Token}'^{(k)}_i$ represents the $k$-th reconstructed token for individual $i$ in the ${t_0}$ input sequence \(
\tilde{R}^{t_0}_i
\). 

This model is then trained exclusively on normal pose sequences, allowing it to learn typical movement patterns and accurately reconstruct inputs with a low Mean Squared Error (MSE) loss. Since the decoder's target sequence is set to match the encoder's input sequence, the MSE loss between the reconstructed sequence $\hat{R}^{t_0}_i$ and the encoder input sequence \(
\tilde{R}^{t_0}_i
\) is used as the training loss and the basis for computing the normality score during evaluation. However, when the model is exposed to shoplifting pose sequences, its ability to reconstruct them deteriorates, leading to a higher MSE loss, which serves as a strong indicator of shoplifting incidents.


%% file: sec/experiments.tex
\section{Experimental Setup}
\label{sec:experiments}
\subsection{Dataset}
To ensure a fair and unbiased assessment of Shopformer, we employed the PeseLift dataset \cite {rashvand2025exploring}, the only existing pose-based specifically designed for shoplifting detection. PoseLift \cite {rashvand2025exploring} provides privacy-preserving pose sequence data that capture human motion and behavior without relying on raw video frames. The dataset includes frame-level annotations, where each frame is labeled as either shoplifting or non-shoplifting (normal) based on observable customer behaviors. Annotations also include frame IDs, person IDs for tracking individuals across frames, and 2D human pose representations using the COCO17 keypoint format to accurately capture body movements. These annotations were generated using a pipeline that integrates YOLOv8 for person detection, ByteTrack for multi-person tracking, and HRNet for accurate pose estimation \cite {rashvand2025exploring}.

The PoseLift \cite {rashvand2025exploring} has been structured by its authors into separate training and testing sets to facilitate standardized and unbiased evaluation. Following the default training-test split provided, we maintain consistency and comparability across all experiments. 
Designed specifically for unsupervised shoplifting detection, the training set contains only normal behaviors commonly observed in retail environments, such as walking, browsing, and picking up items. In contrast, the test set includes both normal and shoplifting behaviors, covering a diverse range of shoplifting scenarios, including concealing items in pockets, hiding them under clothing, and placing them in bags.

\subsection{Metrics}
Selecting appropriate evaluation metrics is essential for effectively measuring model performance in the context of shoplifting detection. Given the inherent class imbalance in this task, where shoplifting incidents are significantly less frequent than normal behaviors, informative evaluation metrics are critical to ensure fair and meaningful assessment. In our experiments, we adopted standard metrics commonly used in anomaly detection with unsupervised learning, including AUC-ROC, AUC-PR, and Equal Error Rate (EER). These metrics provide complementary insights into the models’ ability to distinguish between normal and shoplifting (anomalous) behaviors under imbalanced data conditions. A detailed explanation of each metric is provided in the following subsections.
\subsubsection{AUC-ROC}
The Area Under the Receiver Operating Characteristic Curve (AUC-ROC) is a widely used metric for evaluating performance in anomaly detection. The ROC curve illustrates the model’s ability to distinguish between normal and anomalous instances by plotting the True Positive Rate (TPR) against the False Positive Rate (FPR) across various decision thresholds. A higher AUC-ROC value indicates better discriminative capability. However, it is important to note that AUC-ROC does not account for the False Negative Rate (FNR), which may be critical in applications like shoplifting detection where missing a shoplifting event can have serious consequences. Therefore, relying solely on AUC-ROC may not provide a complete picture of model performance, and it should be complemented with additional metrics for a more thorough evaluation.
\subsubsection{AUC-PR}
The Area Under the Precision-Recall Curve (AUC-PR) is a key performance metric for shoplifting detection tasks involving highly imbalanced datasets. AUC-PR focuses on the model's ability to correctly identify the positive (shoplifting) class. The Precision-Recall (PR) curve plots precision against recall at different decision thresholds, capturing the trade-off between correctly identifying anomalies and minimizing false alarms. A higher AUC-PR score reflects a model's stronger capability in detecting shoplifting (anomalous) events, making it particularly suitable for applications where false negatives are costly.
\subsubsection{Equal Error Rate (EER)}
The Equal Error Rate (EER) is a performance metric that identifies the point where the FPR and FNR are equal. In other words, EER reflects the threshold at which the rate of incorrectly flagging normal behaviors as shoplifting matches the rate of failing to detect actual shoplifting events. A lower EER value indicates a more effective model, as it demonstrates a balanced reduction of both error types. Given that minimizing both false alarms and missed detections is critical in shoplifting detection systems, EER serves as an important and practical metric for assessing model performance under these real-world constraints.
\subsection{Training and evaluation Strategy}

The training set exclusively comprises normal shopping activities, allowing models to learn typical customer behavior without exposure to shoplifting incidents. It includes 53,353 normal frames and no shoplifting frames. To mitigate bias, the test set is designed with a balanced distribution, containing 1,500 shoplifting frames from 43 distinct shoplifting incidents and 2,221 normal frames.  This near-equal distribution ensures a fair assessment of model performance in distinguishing shoplifting behaviors. For consistency and reliability, we used this predefined training-test split across all experiments, ensuring that comparisons between different methods are both fair and reliable.


We evaluated our proposed model against three state-of-the-art pose-sequence-based models developed for anomaly detection. All these methods are designed within an unsupervised learning framework, where models learn to capture normal behavioral patterns in retail environments without access to labeled anomalies during training. Specifically, we compared our approach with STG-NF \cite{hirschorn2023normalizing}, GEPC \cite{markovitz2020graph}, and TSGAD \cite{noghre2024exploratory}, all of which have demonstrated strong performance in detecting abnormal human behavior using pose sequences in the domain of anomaly detection. Although SPARTA \cite{noghre2024posewatch} is a relevant baseline due to its transformer-based architecture, we were unable to include it in our comparison because its codebase is not publicly available. Following the unsupervised learning paradigm, these models are trained to minimize a specific objective function that enables them to model normal customer behavior. In the training phase, we used the default setting from the original papers for each model and our experimental setup was implemented using the PyTorch framework. All experiments were conducted on a GPU server equipped with four NVIDIA Tesla V100 GPUs, each with 32 GB of memory.  After training, they are evaluated on a separate test set containing both normal and shoplifting instances to assess their ability to identify shoplifting events as anomalies. This comparative evaluation allows us to measure the effectiveness of our method relative to established benchmarks in the field.

\subsection{Token number selection: Finding the Optimal Trade-off}
\label{sec:token}
As discussed in \cref{sec:Shopformer}, our tokenizer module leverages the encoder from the trained autoencoder to generate tokens, which serve as input to the transformer. To gain deeper insights into how the number of tokens and the resolution of the feature map influence model performance, we design two complementary sets of experiments.

In the first experiment, we aim to investigate the impact of the number of tokens when the feature map is fixed. To ensure consistency, we keep the feature map produced by the tokenizer unchanged across all trials while setting the window size to 12. We then systematically vary the number of tokens generated by the tokenizer module from 2 to 6 and analyze the corresponding effect on the model's performance. 
This setup allows us to explore different tokenization granularities and identify a suitable number of tokens that effectively capture essential spatial information while balancing computational cost. By fixing the feature map and only adjusting the token count, we aim to assess how much information can be compressed into tokens without degrading the performance of the downstream task. This analysis helps identify an optimal trade-off between model complexity and representational capacity.

In the second experiment, we explore how variations in the feature map itself affect performance when the number of tokens is fixed. The token count is set based on the optimal value identified in the first experiment, and we vary the feature map dimensions generated by the tokenizer module. Specifically, we expand the channels from the initial 2-channel representation, which corresponds to the x and y coordinates of the $K$ keypoints, to richer feature encodings, increasing to 4, 8, 12, 16, 20, 28, 32, 64 channels.
This setup allows us to study how the richness of the feature map, controlled through its spatial resolution and channel depth, influences the quality of the resulting tokens. By fixing the number of tokens and adjusting only the feature map, we aim to analyze how much spatial detail and semantic content need to be preserved at the feature level to ensure effective token representation and transformer processing. 

To ensure consistency across all experiments, we used the same transformer configuration throughout the token ablation studies. Specifically, the transformer module was configured with 12 attention heads, 4 encoder-decoder layers, and a feed-forward dimensionality of 64. This setting was chosen as a balanced baseline, allowing us to isolate and evaluate the impact of token count and feature map resolution without interference from changes in the transformer’s architecture. By keeping the transformer fixed, we ensure that observed performance variations are attributable to tokenization parameters alone, thereby providing a clearer understanding of how different tokenization strategies affect the model's ability to detect shoplifting behaviors.

%% file: sec/results.tex
\section{Results}
\label{sec:results}
In this section, we present the experimental findings that support the effectiveness of Shopformer. We begin by reporting the results of our ablation studies on token number and feature map size (detailed in Section~\ref{sec:token}), which helped identify the best-performing configuration.
Based on extensive experiments in selecting the optimal number of tokens and feature map size, we identify our best-performing model and compare it against the state-of-the-art pose-based anomaly detection methods. 

\begin{figure}[htbp]
  \centering
   \includegraphics[trim=50pt 30pt 40pt 50pt, clip, width=3.4 in]{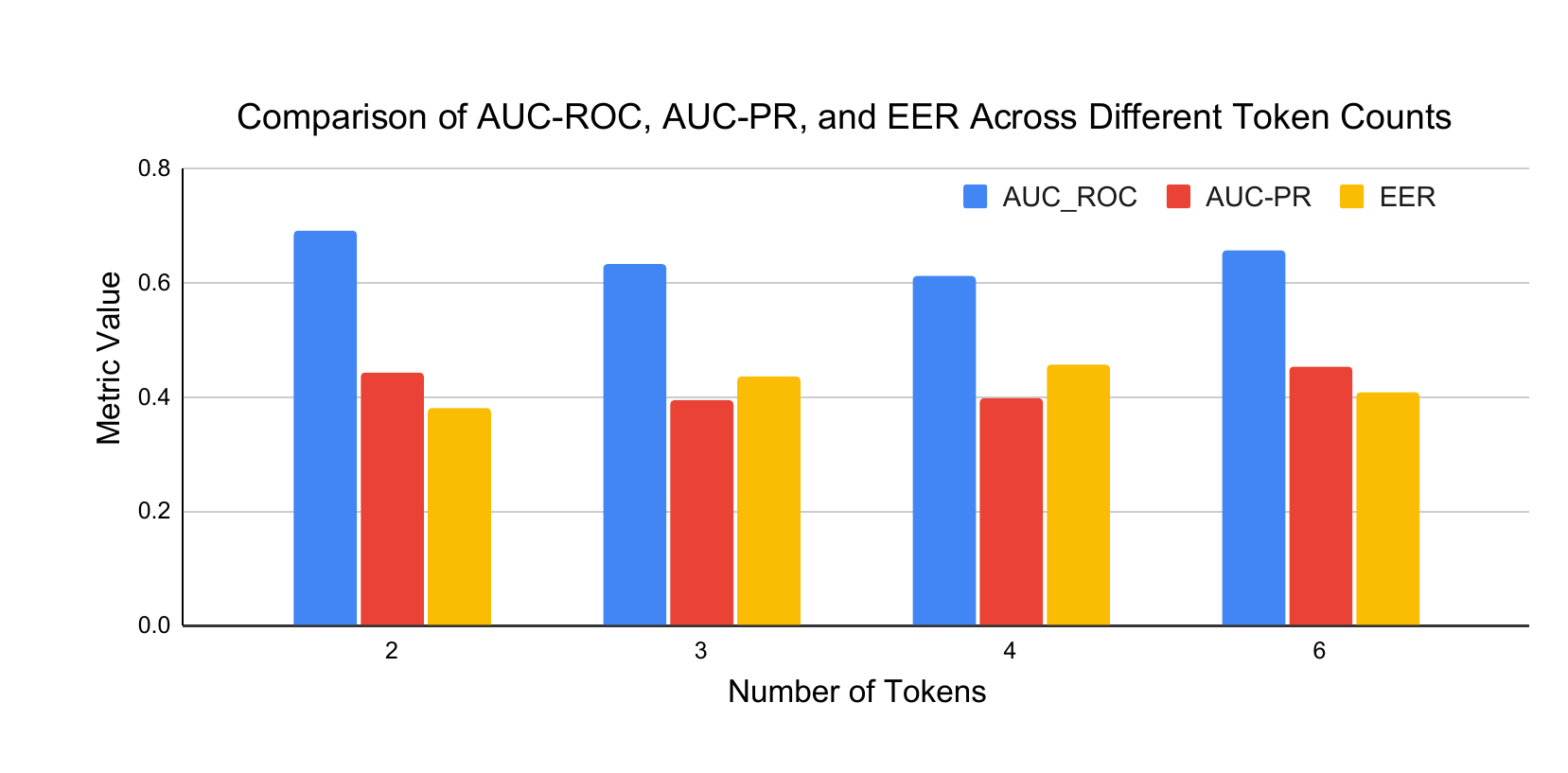}
   \caption{Comparison of AUC-ROC, AUC-PR, and EER across different token counts, highlighting 2 tokens as the optimal choice with the highest AUC-ROC (0.6915) and lowest EER (0.3819).}
   \label{fig:merics_token}
\end{figure}

As shown in \cref{fig:merics_token},
the comparison of AUC-ROC, AUC-PR, and EER, across different token numbers, reveals a non-linear relationship between the number of tokens and model performance, with varying impacts observed across all evaluation metrics. The AUC-ROC, which measures the model’s ability to distinguish between shoplifting and normal behavior, is highest at 2 tokens (69.15\%). However, as the token count increases to 3 and 4, AUC-ROC drops significantly to 63.26\% and 61.17\%, respectively, suggesting that increasing the number of tokens may weaken the meaningful information each one carries. At 6 tokens, AUC-ROC slightly recovers to 65.83\%, but still falls short of the initial performance with 2 tokens.

AUC-PR, which emphasizes the model's precision in detecting the minority class (shoplifting), follows a similar trend. It starts at 44.50\% for 2 tokens, dips to 39.34\% and 39.99\% for 3 and 4 tokens, respectively, and reaches its highest value of 45.28\% at 6 tokens. This indicates that while 6 tokens provide better precision-recall balance, it does not compensate for the drop in AUC-ROC. 
Regarding the Equal Error Rate (EER), which reflects the point where false acceptance and false rejection rates are equal, the lowest error (0.3819) is achieved at 2 tokens. EER worsens with more tokens, peaking at 4 tokens (0.4559), then improves slightly at 6 tokens (0.4097). Overall, while 6 tokens yield the best AUC-PR, the optimal balance between high AUC-ROC and low EER is achieved with 2 tokens.


\cref{tab:feature} presents the results of the second experiment, which examines how variations in feature map size influence performance while keeping the number of tokens fixed at two. In this experiment, the feature map produced by the backbone (tokenizer module) is progressively expanded in channel depth, increasing from an initial 2-channel representation to richer feature encodings with 4, 8, 12, 16, 20, 28, 32, 64 channels, which correspond to token embedding size of 72, 144, 216, 288, 360, 504, 576, and 1152, respectively. This setup allows us to explore the trade-off between richer feature representations and their effectiveness in transformer-based token processing. 
From the table, we observe that increasing the token embedding size, which corresponds to a more complex and higher-capacity feature map, does not consistently lead to improved performance. Notably, the best result for 2 tokens is achieved with a token embedding size of 144, yielding the highest AUC-ROC of 69.15. 

\begin{table}[htpb]  
\centering
\small 
\caption{Comparison of AUC-ROC, AUC-PR, and EER for different token embedding size using 2 tokens generated from the tokenizer.}
\label{tab:feature}
\resizebox{\columnwidth}{!}{%
\begin{tabular}{c|c|c|c|c}
\toprule
\textbf{Number of Tokens} & \textbf{Token Embedding Size} & \textbf{AUC-ROC} & \textbf {AUC-PR} & \textbf{EER} \\ 
\midrule
\multirow{5}{*}{} 
2   & 72 & 62.20 & 40.52 &  0.44 \\
2   & 144 & 69.15 & 44.49 &  0.38 \\
2   & 216 & 62.34 & 39.83 & 0.44  \\
2 & 288 & 62.11 & 40.14 & 0.46 \\
2   & 360 & 61.84 & 39.50 & 0.45   \\
2   & 504& 65.49 & 43.19 &  0.41 \\
2  & 576 & 64.57 &  43.51 & 0.42 \\
2 & 1152 & 68.33 & 45.16 & 0.40 \\
\bottomrule
\end{tabular}
}
\end{table}

 We then compare our best model with two tokens against the state-of-the-art pose-based anomaly detection approaches. Before presenting the final comparison, we conducted ablation studies to analyze the impact of transformer architecture choices on Shopformer's performance. Specifically, we varied the number of layers in the encoder and decoder (\(L=2, 4, 6\)), the number of attention heads \((T=2, 4, 8, 12\)), and the feed-forward dimensionality (\(F=64, 128\)). 
 
 To balance performance with computational efficiency, we adopt a lightweight transformer configuration suitable for real-time inference. Based on the results of our ablation studies, we selected an efficient configuration consisting of 2 transformer layers, 2 attention heads, and a feed-forward dimensionality of 64, used in conjunction with the optimal 2-token input. This configuration offers a strong trade-off between detection accuracy and computational overhead. Detailed results and further analysis of these transformer ablation experiments are provided in the supplementary materials.

 
 \cref {tab:models} presents the performance of Shopformer on the PoseLift dataset \cite{rashvand2025exploring}, compared to the three state-of-the-art pose-based anomaly detection models. The results demonstrate that Shopformer achieves the highest accuracy in shoplifting detection, surpassing the previous state-of-the-art STG-NF model by 67.46\% in AUC-ROC. STG-NF \cite{hirschorn2023normalizing}, which models normal behavior using a fixed probability distribution, struggles to adapt to the variability in shoplifting actions, resulting in a 1.69\% performance drop. In contrast, Shopformer’s transformer-based design offers greater modeling capacity and adaptability, allowing it to more effectively learn diverse behavioral patterns and outperform existing approaches.

\begin{table}[htp]
\centering
\caption{AUC-ROC, AUC-PR, and EER of Shopformer compared with state-of-the-art pose-based anomaly detection models on the PoseLift dataset \cite{rashvand2025exploring}.}
\label{tab:models}
\resizebox{0.94\columnwidth}{!}{%
\begin{tabular}{lc|ccc}
\toprule[\heavyrulewidth] \midrule
\textbf{Methods}     & \textbf{Venue}  & \textbf{AUC-ROC} & \textbf{AUC-PR} & \textbf{EER} \\ \midrule
\textbf{STG-NF \cite{hirschorn2023normalizing}}   & ICCV 2023 & 67.46         & 84.06 & 0.39 \\
\textbf{TSGAD \cite{noghre2024exploratory}}   & WACV 2024 & 63.35      &  39.31  & 0.41 \\
\textbf{GEPC \cite{markovitz2020graph}}   & CVPR 2020 &  60.61  & 50.38   & 0.38  \\
\midrule
\textbf{Shopformer}   & CVPR 2025 & 69.15    & 44.49    & 0.38  \\

\midrule \bottomrule[\heavyrulewidth]
\end{tabular}%
}
\end{table}

%% file: sec/conclusion.tex
\section{Conclusion}
\label{sec:Conclusion}

In this work, we introduced Shopformer, a novel transformer-based model designed specifically for shoplifting detection using human pose sequences. Addressing the limitations of traditional pixel-based methods, including lack of privacy preservation, limited scalability, and potential biases, Shopformer leverages a graph convolutional autoencoder tokenizer and transformer architecture to capture the spatio-temporal dynamics of human behavior in a privacy-preserving and computationally efficient manner for the shoplifting detection task. 

Through comprehensive experiments on the PoseLift dataset, the only available real-world, pose-based benchmark for shoplifting detection, Shopformer demonstrates superior performance over state-of-the-art pose-based anomaly detection models. Our model achieves the highest AUC-ROC score of 69.15\%, highlighting its ability to accurately distinguish between normal and shoplifting behaviors. These results confirm the effectiveness of our tokenization strategy and transformer architecture and validate the potential of transformer-driven, pose-sequence-level models as a privacy-preserving, fair, and real-time solution for shoplifting detection in real-world retail environments.

\section*{Acknowledgment}
This research is supported by the National Science Foundation (NSF) under Award Number 2329816.

%% file: sec/suppl.tex
\clearpage

\section*{Supplementary Material}
\label{sec:sup}
This section provides a comprehensive overview of the ablation studies conducted to determine the optimal configuration of Shopformer. It includes various experimental setups, architectural variations, training strategies, hyperparameter configurations, and detailed results. We report how changes in token count, transformer depth, attention heads, and feature map dimensions affect model performance across multiple evaluation metrics, supporting both reproducibility and deeper analysis of Shopformer's architecture under different configurations.  

\subsection{Ablation Study on Tokenization Parameters and Feature Map Size}
To analyze how different architectural choices affect Shopformer’s performance, we conducted a comprehensive grid-style ablation study targeting different design choices. One key factor was the number of tokens and token embedding size, as detailed in \cref{sec:token}, with a subset of results shown in \cref{fig:merics_token} and \cref{tab:feature}. 

A more comprehensive set of ablation study results is presented here. Specifically, we varied both the number of tokens (1, 2, 3, 4, 6, and 12) and the token embedding sizes (72, 144, 216, and 288), which correspond to the feature map dimensions output by the tokenizer. This allows us to evaluate the trade-offs between token granularity and feature richness in the learned representations. 
For these experiments, we utilized the Shopformer configuration with 12 attention heads, 4 transformer layers, and a feed-forward dimension of 64. All models were trained using the Adam optimizer with a learning rate of $5 \times 10^{-5}$ for 20 epochs. We followed the same two-stage training strategy described in \cref{sec:Shopformer}.

First, the GCAE is trained independently. Once complete, its encoder is frozen and repurposed as the tokenizer module, as illustrated in \cref{fig:shopformer}. The transformer module is then trained on top of this frozen encoder. The full configuration used for these experiments is summarized in Table ~S1. 
\begin{table}[ht]
\captionsetup{labelformat=empty}
\centering
\caption{\textbf{Table S1.} Shopformer training setup, including the hyperparameters used for the token count and feature size ablation study}
\label{tab:shopformer_config}
\resizebox{0.998\linewidth}{!}{%
\begin{tabular}{ll}
\hline
\textbf{Parameter} & \textbf{Value} \\
\hline
Training Strategy            & Two-stage (Autoencoder + Transformer) \\
Epochs                       & 20 \\
Input Frame Size             & 12 \\
Attention Heads              & 12 \\
Transformer Layers           & 4 \\
Feed-Forward Dimension     & 64 \\
Optimizer                    & Adam \\
Learning Rate    &  $5 \times 10^{-5}$\\
Dropout Rate    &  $1 \times 10^{-1}$        \\

\hline
\end{tabular} }
\end{table}

The results of these experiments, reported in Table S2, include standard evaluation metrics such as AUC-ROC, AUC-PR, and EER. Additionally, we report EER Threshold (EER TH), 10\% Error Rate (10 ER), and its corresponding threshold (10 ER TH), which are less commonly used in anomaly detection literature. These values are included in the supplementary materials to provide a more detailed view of model behavior under specific operating conditions. Including these additional metrics also supports reproducibility by allowing other researchers to evaluate the model across a broader range of decision thresholds and performance criteria. A brief explanation of these values follows: the EER TH refers to the decision threshold at which the FPR equals the FNR. 10 ER indicates the FNR when the FPR is fixed at 10\%, and 10 ER TH is the corresponding threshold at which this error rate is measured.

As shown in Table S2, performance varies significantly with token count and token embedding size. Our ablation study shows that using 2 tokens with an embedding size of 144 achieves the best overall results with the selected transformer configuration, reaching the highest AUC-ROC (69.14\%) and the lowest EER (38.19\%). These findings support our design choices, as discussed in \cref{sec:token} with results reported in \cref{{sec:results}}.

Based on these observations, we selected two configurations for subsequent ablation studies. The first is the best-performing setup: 2 tokens with a 144-dimensional embedding. 
The second is a lightweight alternative using 2 tokens with a 72-dimensional embedding, chosen for its lower computational cost and relevance to real-time shoplifting detection applications. While its current performance is limited, we explore whether further tuning of the transformer configuration can improve its effectiveness, as discussed in \cref{transformer ablation}.

\begin{table*}[htpb] 
\captionsetup{labelformat=empty}
\centering
\scriptsize
\caption{\textbf{Table S2.} Performance of Shopformer (AUC-ROC, AUC-PR, EER, EER TH, 10 ER, and 10 ER TH) with token counts of 1, 2, 3, 4, 6, and 12, and token embedding sizes of 72, 144, 216, and 288}
\label{tab:tokens_f}
\resizebox{0.76\textwidth}{!}{%
\begin{tabular}{c|c|c|c|c|c|c|c}
\toprule
\textbf{Number of Tokens} & \textbf{Token Embedding Size} & \textbf{AUC-ROC} & \textbf {AUC-PR} & \textbf{EER} & \textbf{EER TH} & \textbf{10 ER} & \textbf{10 ER TH}\\ 
\midrule
\multirow{5}{*}{} 
1   &  72 &  62.19  &  40.96& 44.26 & 16.37 & 0.69& 1.07\\ 
2   & 72 & 62.22 & 40.52&  43.93 &  63.28& 0.66 & 23.45\\
3  & 72 &  62.37 & 36.55 &  43.39&  24.11& 0.65 & 5.27 \\
4 &  72& 61.20 & 37.30 &  43.33 & 76.24 & 0.68& 7.29\\ 
 6  & 72 & 63.09 & 39.49 &  42.39 &  57.49&   0.64 & 29.57\\   
 12  &72& 60.93  &  35.78& 45.65  &  30.11& 0.67 &  7.99\\   
1   &  144 &  64.79& 42.45 & 43.78 &29.15 & 0.61 & 17.46 \\
2   & 144 & 69.14 &  44.50&  38.19 & 159.10 & 0.60 & 112.97 \\
3  & 144 &  63.26& 39.34 & 43.63  & 73.63 & 0.63 & 39.33\\
4 &  144&  61.17& 39.99 &  45.59 & 28.17 & 0.63 & 12.07\\
 6  & 144 & 65.83 & 45.28 &  40.97 & 117.06 & 0.62 & 52.16  \\
 12  &144& 68.06  & 45.49 &  40.22 & 129.93 & 0.59 & 87.35 \\
1   &  216 & 66.27 &  43.98 & 42.63 & 83.58 & 0.62 & 35.00\\ 
2   & 216 & 62.34 & 39.83&  44.66 &  171.27& 0.65 & 74.59 \\ 
3  & 216 & 63.25 & 39.34 &  43.63& 73.99  & 0.63 & 39.82\\
4 &  216 &  63.59&  41.06&  44.38 & 92.51 & 0.61& 59.75\\ 
 6  & 216 &  61.47&  39.49 &  45.53 & 81.99 &  0.65&  43.64 \\ 
 12  & 216&  62.06 & 40.19 &  44.50 & 72.38  & 0.66 & 35.23 \\ 
1   &  288 & 64.67 &  42.81& 43.21 & 161.96& 0.62 & 76.06 \\ 
2   & 288 & 62.11  & 40.14 &  46.04 &  129.99 & 0.64 & 65.95\\ 
3  & 288 & 64.61 & 41.49  &  41.79&  102.97& 0.66 & 46.67 \\ 
4 &  288 & 64.75 & 44.56  &  43.05 & 141.57 & 0.66& 45.54\\
 6  & 288 & 61.58 &  40.02&  45.38 & 123.62 & 0.65 & 58.77  \\
 12  & 288&  64.11 & 43.07 &  41.70 &  193.79& 0.63 &  110.30\\  
\bottomrule
\end{tabular}
}
\end{table*}
\subsection{Analyzing the Effect of Transformer Configuration}
\label{transformer ablation}

This subsection explores the impact of transformer configuration on the performance of Shopformer, with the token count fixed at two. 
Specifically, we examine how varying the number of attention heads (2, 4, 8, 12), the number of transformer layers (2, 4, 6), and the size of the feed-forward network (64, 128) influence detection accuracy. By systematically adjusting these parameters, we aim to identify configurations that offer the best trade-off between model complexity and performance. As shown in Table S3, adjusting the transformer configuration did not lead to a noticeable improvement in accuracy for the 72-dimensional token embeddings, with AUC-ROC consistently around 62\%. Based on the configuration using 2 tokens with a 144-dimensional embedding, we selected the optimal model: 2 transformer layers, 2 attention heads, and a feed-forward dimension of 64. This setup achieves high accuracy while maintaining a lower parameter count compared to other configurations.

\begin{table*}[htpb]  
\centering
\small 
\caption*{\textbf{Table S3.} Transformer configuration ablation study on Shopformer using 2-token input sequences.  The table shows the impact of varying the number of transformer layers, attention heads, and feed-forward dimensions on detection performance, evaluated using AUC-ROC, AUC-PR, EER, EER TH, 10 ER, 10 ER TH. Experiments are conducted using two feature sizes, 72 and 144, corresponding to 4 and 8 output channels from the GCAE encoder, respectively. The best-performing configuration, considering both accuracy and model complexity, is highlighted in bold.
}
\label{tab:tokens_2}
\resizebox{0.93\textwidth}{!}{%
\begin{tabular}{c|c|c|c|c|c|c|c|c|c}
\toprule
\textbf{Layers} & \textbf{Attention Heads} & \textbf{Feed-forward Dimension} & \textbf{Token Embedding Size} &\textbf {AUC-ROC} & \textbf{AUC-PR} & \textbf{EER} & \textbf{EER TH} & \textbf{10 ER} & \textbf{10 ER TH}  \\ 
\midrule
\multirow{5}{*}{} 
2 & 2  & 64 & 72 &  62.22& 40.52 & 43.96& 59.73 & 0.66 & 21.21 \\
2 & 2  & 128 & 72 & 62.21 &  40.51&  43.99 & 59.84 & 0.66 & 21.37\\
  2 & 4  & 64 & 72 &  62.21&  40.52& 43.93 & 59.78 & 0.66& 21.18\\
  2 & 4  & 128 &  72& 62.20 & 40.53& 44.05 & 56.01 & 0.66  & 18.62\\
   2 & 8 & 64 & 72 & 62.22 & 40.54& 43.93& 59.54 & 0.66 & 21.06\\
 2 & 8 & 128 & 72 & 62.21 & 40.51& 43.99& 59.37 &  0.66 &21.30\\
  2 & 12 & 64 & 72 &  62.22& 40.50& 44.02 & 59.81& 0.66  &  21.39\\
   2 & 12 & 128 & 72  & 62.21 & 40.52 & 44.02& 59.81 & 0.66  & 21.39\\
4 &  12& 64 & 72 & 62.22& 40.52& 43.93 &  63.28 &  0.66& 23.45\\ 
4 &  12& 128 & 72 & 62.20 & 40.52&  44.08&  56.67&  0.66 & 19.29\\  
 6  & 8 & 64 & 72  & 62.20 & 40.53& 44.08&  57.36&    0.66& 19.90\\ 
\textbf{2} & \textbf{2}  & \textbf{64} & \textbf{144} &  \textbf{69.15} & \textbf{44.49} & \textbf{38.16} & \textbf{156.74} &  \textbf{0.60} & \textbf{111.18}\\   
2 & 2  & 128 & 144 & 69.16  & 44.50 & 38.19&  161.67 & 0.60 & 114.84 \\  
2 & 4  & 64 & 144 & 69.15 &  44.47 & 38.16 &  156.67 & 0.59 & 111.13\\  
2 & 4  & 128 &  144& 69.17 & 44.52& 38.19& 161.99 & 0.60 & 115.11\\
2 & 8 & 64 & 144 & 69.14 &  44.47 & 38.16 & 156.72& 0.59  & 111.15\\ 
 2 & 8 & 128 & 144 & 69.17 & 44.51 &  38.19 & 162.06 & 0.60 & 115.26\\ 
2 & 12 & 64 & 144 & 69.17 &  44.51 & 38.19 & 161.98 & 0.60&  115.11\\  
2 & 12 & 128 &144  & 69.17 & 44.51& 38.19 & 161.98 & 0.60  & 115.11\\ 
4 & 2 & 64 & 144 &  69.14 & 44.50 & 38.19&  160.43&  0.60 & 113.99\\
4 & 2 & 128 & 144 & 69.22  & 44.59& 38.16 & 168.24&  0.59  & 120.83\\ 
4 & 4 & 64 & 144 &  69.14 & 44.48& 38.19 & 157.59 &   0.60& 111.73 \\ 
4 & 4 & 128 & 144 & 69.23  &  44.59& 38.22& 168.60 & 0.60   & 121.05 \\ 
4 & 8 & 64 & 144 & 69.15 & 44.52 & 38.19& 162.01 &  0.60& 115.19 \\
4 &  8& 128 & 144 &  69.21 & 44.56 & 38.25 & 168.46 &  0.60  & 120.93\\
4 &  12& 64 & 144 & 69.14&  44.50& 38.19 & 159.10 &  0.60 & 112.97\\ 
4 &  12& 128 & 144 & 69.21 &  44.47& 38.22 & 168.44&  0.60 & 120.73 \\   
6  & 8 & 64 & 144  & 69.17 & 44.53& 38.19  & 166.75 &  0.60&  118.99\\  
6  & 8 & 128 & 144  &  69.23  & 44.59 & 38.19& 168.74 & 0.60 & 121.03 \\

\bottomrule
\end{tabular}
}
\end{table*}

%% file: main.bbl
\begin{thebibliography}{39}
\providecommand{\natexlab}[1]{#1}
\providecommand{\url}[1]{\texttt{#1}}
\expandafter\ifx\csname urlstyle\endcsname\relax
  \providecommand{\doi}[1]{doi: #1}\else
  \providecommand{\doi}{doi: \begingroup \urlstyle{rm}\Url}\fi

\bibitem[Abshari and Sridhar(2025)]{abshari2025survey}
Danial Abshari and Meera Sridhar.
\newblock A survey of anomaly detection in cyber-physical systems.
\newblock \emph{arXiv preprint arXiv:2502.13256}, 2025.

\bibitem[Alinezhad~Noghre et~al.(2023)Alinezhad~Noghre, Katariya, Danesh~Pazho, Neff, and Tabkhi]{alinezhad2023pishgu}
Ghazal Alinezhad~Noghre, Vinit Katariya, Armin Danesh~Pazho, Christopher Neff, and Hamed Tabkhi.
\newblock Pishgu: Universal path prediction network architecture for real-time cyber-physical edge systems.
\newblock In \emph{Proceedings of the ACM/IEEE 14th International Conference on Cyber-Physical Systems (with CPS-IoT Week 2023)}, pages 88--97, 2023.

\bibitem[Ansari and Singh(2022)]{ansari2022expert}
Mohd~Aquib Ansari and Dushyant~Kumar Singh.
\newblock An expert video surveillance system to identify and mitigate shoplifting in megastores.
\newblock \emph{Multimedia Tools and Applications}, 81\penalty0 (16):\penalty0 22497--22525, 2022.

\bibitem[Ansari and Singh(2023)]{ansari2023optimized}
Mohd~Aquib Ansari and Dushyant~Kumar Singh.
\newblock Optimized parameter tuning in a recurrent learning process for shoplifting activity classification.
\newblock \emph{Cybernetics and Information Technologies}, 23\penalty0 (1):\penalty0 141--160, 2023.

\bibitem[Ardabili et~al.(2022)Ardabili, Pazho, Noghre, Neff, Ravindran, and Tabkhi]{ardabili2022understanding}
Babak~Rahimi Ardabili, Armin~Danesh Pazho, Ghazal~Alinezhad Noghre, Christopher Neff, Arun Ravindran, and Hamed Tabkhi.
\newblock Understanding ethics, privacy, and regulations in smart video surveillance for public safety.
\newblock \emph{arXiv preprint arXiv:2212.12936}, 2022.

\bibitem[Ardabili et~al.(2023)Ardabili, Pazho, Noghre, Neff, Bhaskararayuni, Ravindran, Reid, and Tabkhi]{ardabili2023understanding}
Babak~Rahimi Ardabili, Armin~Danesh Pazho, Ghazal~Alinezhad Noghre, Christopher Neff, Sai~Datta Bhaskararayuni, Arun Ravindran, Shannon Reid, and Hamed Tabkhi.
\newblock Understanding policy and technical aspects of ai-enabled smart video surveillance to address public safety.
\newblock \emph{Computational Urban Science}, 3\penalty0 (1):\penalty0 21, 2023.

\bibitem[Ardabili et~al.(2024)Ardabili, Pazho, Noghre, Katariya, Hull, Reid, and Tabkhi]{ardabili2024exploring}
Babak~Rahimi Ardabili, Armin~Danesh Pazho, Ghazal~Alinezhad Noghre, Vinit Katariya, Gordon Hull, Shannon Reid, and Hamed Tabkhi.
\newblock Exploring public's perception of safety and video surveillance technology: A survey approach.
\newblock \emph{Technology in Society}, 78:\penalty0 102641, 2024.

\bibitem[Arnold et~al.(2024)Arnold, Schiff, Schiff, Love, Melot, Singh, Jenkins, Lin, Pilz, Enweareazu, et~al.]{arnold2024introducing}
Zachary Arnold, Daniel~S Schiff, Kaylyn~Jackson Schiff, Brian Love, Jennifer Melot, Neha Singh, Lindsay Jenkins, Ashley Lin, Konstantin Pilz, Ogadinma Enweareazu, et~al.
\newblock Introducing the ai governance and regulatory archive (agora): An analytic infrastructure for navigating the emerging ai governance landscape.
\newblock In \emph{Proceedings of the AAAI/ACM Conference on AI, Ethics, and Society}, pages 39--48, 2024.

\bibitem[Arroyo et~al.(2015)Arroyo, Yebes, Bergasa, Daza, and Almaz{\'a}n]{arroyo2015expert}
Roberto Arroyo, J~Javier Yebes, Luis~M Bergasa, Iv{\'a}n~G Daza, and Javier Almaz{\'a}n.
\newblock Expert video-surveillance system for real-time detection of suspicious behaviors in shopping malls.
\newblock \emph{Expert systems with Applications}, 42\penalty0 (21):\penalty0 7991--8005, 2015.

\bibitem[Babaey and Faragardi(2025)]{babaey2025detecting}
Vahid Babaey and Hamid~Reza Faragardi.
\newblock Detecting zero-day web attacks using one-class ensemble classifiers.
\newblock 2025.

\bibitem[Brooks et~al.(2024)Brooks, Kim, Opara, Keltner, Fang, Monroy, Corona, Tzirakis, Baird, Metrick, et~al.]{brooks2024deep}
Jeffrey~A Brooks, Lauren Kim, Michael Opara, Dacher Keltner, Xia Fang, Maria Monroy, Rebecca Corona, Panagiotis Tzirakis, Alice Baird, Jacob Metrick, et~al.
\newblock Deep learning reveals what facial expressions mean to people in different cultures.
\newblock \emph{Iscience}, 27\penalty0 (3), 2024.

\bibitem[{Capital One Shopping}(2025)]{capitalone2025shoplifting}
{Capital One Shopping}.
\newblock Shoplifting statistics, 2025.
\newblock Accessed: March 17, 2025.

\bibitem[Chen et~al.(2021)Chen, Zhang, Yuan, Li, Deng, and Hu]{chen2021channel}
Yuxin Chen, Ziqi Zhang, Chunfeng Yuan, Bing Li, Ying Deng, and Weiming Hu.
\newblock Channel-wise topology refinement graph convolution for skeleton-based action recognition.
\newblock In \emph{Proceedings of the IEEE/CVF international conference on computer vision}, pages 13359--13368, 2021.

\bibitem[Connolly et~al.(2012)Connolly, Granger, and Sabourin]{connolly2012adaptive}
Jean-Fran{\c{c}}ois Connolly, Eric Granger, and Robert Sabourin.
\newblock An adaptive classification system for video-based face recognition.
\newblock \emph{Information Sciences}, 192:\penalty0 50--70, 2012.

\bibitem[Gim et~al.(2020)Gim, Lee, Kim, Park, and Nasridinov]{gim2020automatic}
U-Ju Gim, Jae-Jun Lee, Jeong-Hun Kim, Young-Ho Park, and Aziz Nasridinov.
\newblock An automatic shoplifting detection from surveillance videos (student abstract).
\newblock In \emph{Proceedings of the AAAI Conference on Artificial Intelligence}, pages 13795--13796, 2020.

\bibitem[Hirschorn and Avidan(2023)]{hirschorn2023normalizing}
Or Hirschorn and Shai Avidan.
\newblock Normalizing flows for human pose anomaly detection.
\newblock In \emph{Proceedings of the IEEE/CVF International Conference on Computer Vision}, pages 13545--13554, 2023.

\bibitem[Hu et~al.(2025)Hu, Kollias, Papadopoulou, Tzouveli, Wei, and Yang]{hu2025rethinking}
Guanyu Hu, Dimitrios Kollias, Eleni Papadopoulou, Paraskevi Tzouveli, Jie Wei, and Xinyu Yang.
\newblock Rethinking affect analysis: A protocol for ensuring fairness and consistency.
\newblock \emph{IEEE Transactions on Biometrics, Behavior, and Identity Science}, 2025.

\bibitem[Kirichenko et~al.(2022)Kirichenko, Radivilova, Sydorenko, and Yakovlev]{kirichenko2022detection}
Lyudmyla Kirichenko, Tamara Radivilova, Bohdan Sydorenko, and Sergiy Yakovlev.
\newblock Detection of shoplifting on video using a hybrid network.
\newblock \emph{Computation}, 10\penalty0 (11):\penalty0 199, 2022.

\bibitem[Latkowski(2024)]{latkowski2024facing}
Thomas~Christopher Latkowski.
\newblock Facing facts: The effect of facial recognition bans on policing effectiveness.
\newblock Master's thesis, Georgetown University, 2024.

\bibitem[Markovitz et~al.(2020)Markovitz, Sharir, Friedman, Zelnik-Manor, and Avidan]{markovitz2020graph}
Amir Markovitz, Gilad Sharir, Itamar Friedman, Lihi Zelnik-Manor, and Shai Avidan.
\newblock Graph embedded pose clustering for anomaly detection.
\newblock In \emph{Proceedings of the IEEE/CVF Conference on Computer Vision and Pattern Recognition}, pages 10539--10547, 2020.

\bibitem[Mathias et~al.(2014)Mathias, Benenson, Pedersoli, and Van~Gool]{mathias2014face}
Markus Mathias, Rodrigo Benenson, Marco Pedersoli, and Luc Van~Gool.
\newblock Face detection without bells and whistles.
\newblock In \emph{European conference on computer vision}, pages 720--735. Springer, 2014.

\bibitem[Muneer et~al.(2023)Muneer, Saddique, Habib, and Mohamed]{muneer2023shoplifting}
Iqra Muneer, Mubbashar Saddique, Zulfiqar Habib, and Heba~G Mohamed.
\newblock Shoplifting detection using hybrid neural network cnn-bilsmt and development of benchmark dataset.
\newblock \emph{Applied Sciences}, 13\penalty0 (14):\penalty0 8341, 2023.

\bibitem[Nazir et~al.(2023)Nazir, Mitra, Sulieman, and Kamalov]{nazir2023suspicious}
Amril Nazir, Rohan Mitra, Hana Sulieman, and Firuz Kamalov.
\newblock Suspicious behavior detection with temporal feature extraction and time-series classification for shoplifting crime prevention.
\newblock \emph{Sensors}, 23\penalty0 (13):\penalty0 5811, 2023.

\bibitem[Noghre et~al.(2024)Noghre, Pazho, and Tabkhi]{noghre2024exploratory}
Ghazal~Alinezhad Noghre, Armin~Danesh Pazho, and Hamed Tabkhi.
\newblock An exploratory study on human-centric video anomaly detection through variational autoencoders and trajectory prediction.
\newblock In \emph{Proceedings of the IEEE/CVF Winter Conference on Applications of Computer Vision}, pages 995--1004, 2024.

\bibitem[Noghre et~al.(2025)Noghre, Pazho, and Tabkhi]{noghre2024posewatch}
Ghazal~Alinezhad Noghre, Armin~Danesh Pazho, and Hamed Tabkhi.
\newblock Human-centric video anomaly detection through spatio-temporal pose tokenization and transformer, 2025.

\bibitem[Papaioannou et~al.(2022)Papaioannou, Gecer, Cheng, Chrysos, Deng, Fotiadou, Kampouris, Kollias, Moschoglou, Songsri-In, et~al.]{papaioannou2022mimicme}
Athanasios Papaioannou, Baris Gecer, Shiyang Cheng, Grigorios Chrysos, Jiankang Deng, Eftychia Fotiadou, Christos Kampouris, Dimitrios Kollias, Stylianos Moschoglou, Kritaphat Songsri-In, et~al.
\newblock Mimicme: A large scale diverse 4d database for facial expression analysis.
\newblock In \emph{European Conference on Computer Vision}, pages 467--484. Springer, 2022.

\bibitem[Pazho et~al.(2023)Pazho, Noghre, Purkayastha, Vempati, Martin, and Tabkhi]{pazho2023survey}
Armin~Danesh Pazho, Ghazal~Alinezhad Noghre, Arnab~A Purkayastha, Jagannadh Vempati, Otto Martin, and Hamed Tabkhi.
\newblock A survey of graph-based deep learning for anomaly detection in distributed systems.
\newblock \emph{IEEE Transactions on Knowledge and Data Engineering}, 36\penalty0 (1):\penalty0 1--20, 2023.

\bibitem[Pazho et~al.(2024)Pazho, Noghre, Katariya, and Tabkhi]{pazho2024vt}
Armin~Danesh Pazho, Ghazal~Alinezhad Noghre, Vinit Katariya, and Hamed Tabkhi.
\newblock Vt-former: An exploratory study on vehicle trajectory prediction for highway surveillance through graph isomorphism and transformer.
\newblock In \emph{Proceedings of the IEEE/CVF Conference on Computer Vision and Pattern Recognition}, pages 5651--5662, 2024.

\bibitem[Qandeel(2024)]{qandeel2024facial}
Mais Qandeel.
\newblock Facial recognition technology: regulations, rights and the rule of law.
\newblock \emph{Frontiers in big Data}, 7:\penalty0 1354659, 2024.

\bibitem[Rashvand et~al.(2025)Rashvand, Noghre, Pazho, Yao, and Tabkhi]{rashvand2025exploring}
Narges Rashvand, Ghazal~Alinezhad Noghre, Armin~Danesh Pazho, Shanle Yao, and Hamed Tabkhi.
\newblock Exploring pose-based anomaly detection for retail security: A real-world shoplifting dataset and benchmark.
\newblock \emph{arXiv preprint arXiv:2501.06591}, 2025.

\bibitem[Salzmann et~al.(2020)Salzmann, Ivanovic, Chakravarty, and Pavone]{salzmann2020trajectron++}
Tim Salzmann, Boris Ivanovic, Punarjay Chakravarty, and Marco Pavone.
\newblock Trajectron++: Dynamically-feasible trajectory forecasting with heterogeneous data.
\newblock In \emph{Computer Vision--ECCV 2020: 16th European Conference, Glasgow, UK, August 23--28, 2020, Proceedings, Part XVIII 16}, pages 683--700. Springer, 2020.

\bibitem[Sultani et~al.(2018)Sultani, Chen, and Shah]{sultani2018real}
Waqas Sultani, Chen Chen, and Mubarak Shah.
\newblock Real-world anomaly detection in surveillance videos.
\newblock In \emph{Proceedings of the IEEE conference on computer vision and pattern recognition}, pages 6479--6488, 2018.

\bibitem[{U.S. Census Bureau}()]{uscensus}
{U.S. Census Bureau}.
\newblock Census data, population.
\newblock \url{https://data.census.gov/}.
\newblock Accessed: March 20, 2025.

\bibitem[{U.S. Congress}(2023)]{congress2023frt}
{U.S. Congress}.
\newblock Facial recognition and biometric technology moratorium act of 2023, 2023.
\newblock Accessed: 2025-03-21.

\bibitem[{U.S. Government Accountability Office}(2024)]{gao2024frt}
{U.S. Government Accountability Office}.
\newblock Facial recognition technology: Privacy and accuracy issues related to commercial uses, 2024.
\newblock Accessed: 2025-03-21.

\bibitem[Vaswani et~al.(2017)Vaswani, Shazeer, Parmar, Uszkoreit, Jones, Gomez, Kaiser, and Polosukhin]{vaswani2017attention}
Ashish Vaswani, Noam Shazeer, Niki Parmar, Jakob Uszkoreit, Llion Jones, Aidan~N Gomez, {\L}ukasz Kaiser, and Illia Polosukhin.
\newblock Attention is all you need.
\newblock \emph{Advances in neural information processing systems}, 30, 2017.

\bibitem[Wang et~al.(2024)Wang, Wu, Zhou, and Fu]{wang2024beyond}
Xukang Wang, Ying~Cheng Wu, Mengjie Zhou, and Hongpeng Fu.
\newblock Beyond surveillance: privacy, ethics, and regulations in face recognition technology.
\newblock \emph{Frontiers in big data}, 7:\penalty0 1337465, 2024.

\bibitem[Yan et~al.(2018)Yan, Xiong, and Lin]{yan2018spatial}
Sijie Yan, Yuanjun Xiong, and Dahua Lin.
\newblock Spatial temporal graph convolutional networks for skeleton-based action recognition.
\newblock In \emph{Proceedings of the AAAI conference on artificial intelligence}, 2018.

\bibitem[Yu et~al.(2017)Yu, Yin, and Zhu]{yu2017spatio}
Bing Yu, Haoteng Yin, and Zhanxing Zhu.
\newblock Spatio-temporal graph convolutional networks: A deep learning framework for traffic forecasting.
\newblock \emph{arXiv preprint arXiv:1709.04875}, 2017.

\end{thebibliography}
